This version of the article has been accepted for publication, after peer review (when applicable) but is not the Version of Record and does not reflect post-acceptance improvements, or any corrections. The Version of Record is available online at springnature. Use of this Accepted Version is subject to the publisher's Accepted Manuscript terms of use https://www.springernature.com/gp/open-research/policies/accepted-manuscript-terms.

# City-Scale Multi-Camera Vehicle Tracking System with Improved Self-Supervised Camera Link Model

Yuqiang Lin
Department of Mechanical Engineering
University of Bath
Bath, United Kingdom
Yl4300@bath.ac.uk

Sam Lockyer
Department of Mechanical Engineering
University of Bath
Bath, United Kingdom
Sl2726@bath.ac.uk

Nic Zhang
Department of Mechanical Engineering
University of Bath
Bath, United Kingdom
Qz254@bath.ac.uk

**Abstract**. Multi-Target Multi-Camera Tracking (MTMCT) has broad applications and forms the basis for numerous future city-wide systems (e.g. traffic management, crash detection, etc.). However, the challenge of matching vehicle trajectories across different cameras based solely on feature extraction poses significant difficulties. This article introduces an innovative multi-camera vehicle tracking system that utilizes a self-supervised camera link model. In contrast to related works that rely on manual spatial-temporal annotations, our model automatically extracts crucial multi-camera relationships for vehicle matching. The camera link is established through a pre-matching process that evaluates feature similarities, pair numbers, and time variance for high-quality tracks. This process calculates the probability of spatial linkage for all camera combinations, selecting the highest scoring pairs to create camera links. Our approach significantly improves deployment times by eliminating the need for human annotation, offering substantial improvements in efficiency and cost-effectiveness when it comes to real-world application. This pairing process supports cross camera matching by setting spatial-temporal constraints, reducing the searching space for potential vehicle matches. According to our experimental results, the proposed method achieves a new state-of-the-art among automatic camera-link based methods in CityFlow V2 benchmarks with 61.07% IDF1 Score.

## 1. Introduction

With the continuous expansion of urbanisation, city traffic management is becoming more and more challenging. This has lead to increase demand for the development of intelligent transportation systems within the broader concept of smart cities. One of the key enabling elements of those systems is Multi-Target Multi-Camera Tracking (MTMCT), which aims to track vehicles over large areas in networks comprised of multiple traffic cameras. The MTMCT can usually be divided into two main computational stages: single camera tracking (SCT) and inter-camera tracking (ICT). Over recent decades, many researchers have looked into the SCT problem, generating increasingly robust tracking algorithms [1, 2, 3]. Building on the foundations of these SCT algorithms, this paper focuses on the ICT problem.

Despite promising results being achieved in the MTMCT field [18], there are still several challenges making multi-target tracking a difficult task [4]. The first challenge is the *Dramatic Appearance Variabilities:* The most popular methodology for matching tracklets—proto-tracks derived from SCT that include BBox information and concentrated features—across different cameras is by comparing the features of different tracklets. Features such as the edges, centroid, stereo disparity, texture, motion field, and gradients are often retrieved for association [5]. However, variabilities caused by factors like occlusion, changes in illumination, shadows, change in viewing angle and non-rigid deformation can significantly alter the appearance of the same object. In addition to the appearance variabilities, the different *Camera Overlapping Ratios, based on their fields of View (FOV)* can also make it difficult to create robust algorithms adaptable to various scenes. There are three different scenarios: overlapping FOV, mixed FOV and non-overlapping FOV. Both the overlapping FOV and mixed FOV can face the problem of the different overlapping ratios. The level of FOVs overlap affects the formulation of tracking problems in multi-camera systems, thereby affecting the algorithm performance. Furthermore, the *Unknown Number of Targets and Cameras* also complicates the matching process, as the total number of targets is indeterminate, adding complexity to establishing the complete trajectory of each target [6]. This issue becomes even more complex when a tracker erroneously produces multiple local tracklets for a single target within a single camera [7]. In addition to the above three challenges, there are some other challenges, such as *Camera Motion* [8] and *Varying Object Motion* [5] that further contribute to the difficulties faced in MTMCT.

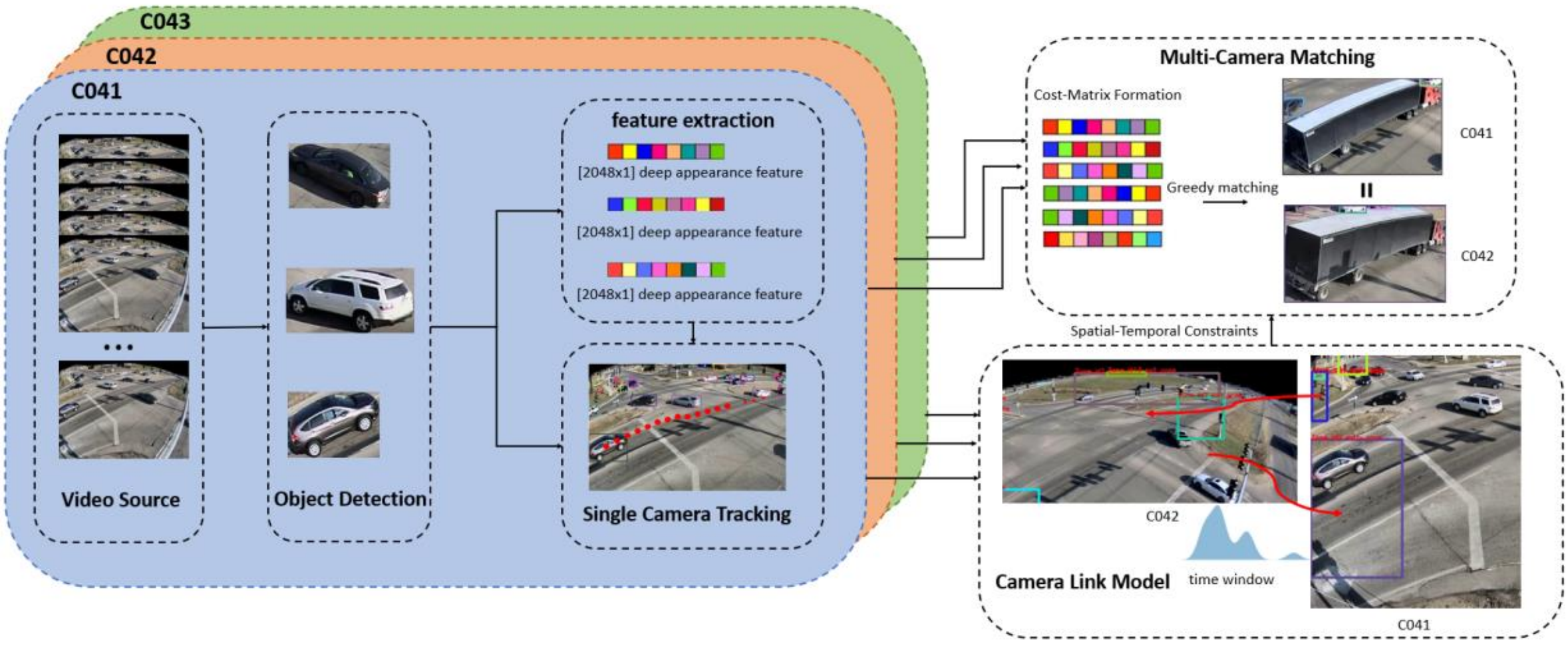

**Figure 1.** The pipeline diagram of our MCVT system.

As it is difficult for multi-camera tracklet matching to deal with these problems in a systematic way, many works have focused on using various techniques [9, 10], including the use of spatial-temporal constraints, zone-based methods, and multimodal information, that aim to reduce the searching space for potential tracklet matches. This reduction minimizes mismatches among visually similar vehicles, thus improving system performance. Among these techniques, spatial-temporal factors are often prioritized for improving the overall performance of MTMCT algorithms, since they have the greatest impact on the searching space. Previous work has manually segmented traffic camera scenes into entry and exit zones, and defined the transition time between cameras [11, 12]. Although the effectiveness of these constraints has been shown [11], they require manual input. This approach may be practical for research studies, with limited camera number and simple camera network settings, but such human-labelling is impractical for real-world applications which. often involve more complex camera networks, variable transition times or temporal road topology changes which require regular updating. Real-world application would thus pose a significant challenge in terms of human resources. To mitigate the demand for manual labor, an automated method known as the camera link model (CLM) has been developed [15] (initially for multi-camera tracking on people). Later, with the rising popularity of the multi camera vehicle tracking (MCVT) problem, CLM was adapted to MCVT [14, 16, 17]; however, these adaptations still required some human labeling and focused primarily on overlapping and mixed FOVs. To the best of our knowledge, [18] is the only work that applies a self-supervised CLM capable of handling non-overlapping FOV camera networks, but this self-supervised CLM still has a focus on overlapping and mixed FOV.

To address this issue, this paper proposes an improved self-supervised CLM focusing on Non-overlapping FOV capable of automatically generating and updating spatial-temporal constraints for MCVT. Compared with the SOTA self-supervised CLM, our proposed CLM introduces a direction vector based pre-matching solution to achieve better zone generation. Additionally, it introduces an evaluation score system to improve zone pairing and integrates kernel density probability into the cost matrix for more precise time constraints. These enhancements improve the overall performance compared to previous CLMs, making it better suited for Non-overlapping FOV.

The pipeline diagram of our proposed system is shown in Figure 1, where each colored box represents a single camera view. Initially, each video frame is processed using a YOLO [19]-based one-stage object detector. The detection outputs are then utilized for extracting deep appearance features. These deep embeddings, combined with the object detection results, are used for a DeepSort [1] based SCT. The tracks obtained from the single camera tracking phase are then used to train the self-supervised CLM. Finally, by utilizing the spatial-temporal constraints generated by the CLM together with the results from single camera tracking, the minimal distances among cross-camera track pairs are identified. These paired tracks are then assigned the same track ID, ensuring consistent identification across multiple cameras.

The rest of this article is structured as follow: an overview for related works is presented in section 2, followed by the methodology in section 3. The experiment results are present in section 4 and, finally, discussion and conclusions are given in section 5.

## 2. Related Works

With the aim of linking the same objects captured by different cameras, object re-identification (Re-ID) has attracted growing attention over recent years. In general, the Re-ID algorithm calculates the similarity between each query-gallery pair and finds the best potential matched targets. Based on the input data formats, object Re-ID methodologies can be divided into two types: image-based approaches and video-based approaches. Theoretically, video-based Re-ID approaches should out-perform image-based Re-ID due to their additional information. Apart from the features extracted from every single frame, the time-related correlations (i.e., spatial–temporal information, trajectory features, vehicle transition order) are also useful in the cross-camera Re-ID comparison [20]. Early research [21, 22, 23, 24] focussed more on imaged-based object Re-ID due to the limitations of SCT algorithm performance and computational resources. The maturity of SCT algorithms [1, 2, 3] coupled with hardware developments have given new impetus to the research into video-based Re-ID [36], hence giving new insights to the MTMCT problem.

The MCVT problem has garnered increased attention following the release of the open-source dataset -- CityFlow [25] on AICITY challenge [26]. To the best of our knowledge, CityFlow is the only real-world open-source dataset specifically for MCVT problem. While there are also some other open-source MCVT synthetic dataset e.g. Synthehicle [27], most MCVT research is conducted on the real-world CityFlow dataset on which our review is focused. Following research developments over 2020-2021, a significant performance improvement resulted from the involvement of human-labelled spatial-temporal constraints as illustrated in Table 1. According to [11], the use of spatial-temporal constraints contributed to a 0.2346 increase in the IDF1 score [35], highlighting the effectiveness of this approach. However, while human labelling proves beneficial for datasets like CityFlow, it is still not applicable for real-world MCVT due to the much more complex and variable parameters that need to be considered. This has motivated researchers to investigate the use of CLM to automatically generate spatial-temporal constraints. CLM was initially applied to non-overlapping FOV scenarios for multi-camera people tracking [31, 32] and then extended to MCVT [14]; however, this CLM work also requires human trajectory annotation so is not fully automated. More recent approaches introduced an improved version of CLM that can autonomously segment entry/exit zones [16, 17], although they continued to rely on human-annotated trajectory data to establish zone connections.

Finally, [18] presents the first attempt to remove the need for human labeling by proposing a self-supervised CLM model that provides the spatial-temporal constraints without the demand for human annotation. The self-supervised CLM shows great potential in real-world application and achieves the SOTA performance on overlapped and mixed FOV CityFlow V1 test scenes [16, 17]. However, despite the advancements, CLM based methods still cannot compete with human annotated approaches, when purely considering accuracy, on non-overlapping FOV CityFlow V2 test scenes [18]. This is due to data and algorithm limitations of CLM itself, which produce less perfect zoning and time transition estimation than human annotations.

**Table 1.** Representative Multi Camera tracking research using the CityFlow dataset. HL denotes Human-labelled, IDF1 is the evaluation metric [35], main methodology briefly illustrates the key methodology followed by the pipeline of object detection, single camera tracking, feature extraction and multi-camera matching.

| Reference | Year | Main Methodology | IDF1 Score | Constraint types |
|---|---|---|---|---|
| City-Scale Multi-Camera Vehicle Tracking by Semantic Attribute Parsing and Cross-Camera Tracklet Matching [28] | 2020 | DCNN based detector with graph clustering for local tracklet generation + ResNet feature extraction +Tracklet-to-Target Assignment (TRACTA) algorithm with traffic topology reasoning component. | 0.4400 | HL spatial constraint |
| Electricity: An efficient multi-camera vehicle tracking system for intelligent city [29] | 2020 | Mask R-CNN for object detection + DeepSort + ResNet feature extraction + Cost matrix solving. | 0.4585 | N.A. |
| A robust MTMC tracking system for ai-city challenge 2021 [30] | 2021 | Cascade R-CNN for detection + TPM tracking + CNN based deep representation + cost matrix solving with spatial-temporal constraints. | 0.7787 | HL spatial-temporal constraint |
| City-scale multi-camera vehicle tracking guided by crossroad zones [11] | 2021 | YOLOv5 + DeepSort based SCT + ResNet feature extraction + zone based spatial-temporal cost-matrix solving. | 0.8095 | HL spatial-temporal constraint |
| Multi-camera vehicle tracking system for AI City Challenge 2022 [12] | 2022 | YOLOv5 + (DeepSort based framework + MedianFlow + Efficient Convolution Operator for SCT) + ResNet based deep feature extraction + cost matrix solving with spatial-temporal constraints. | 0.8437 | HL spatial-temporal constraint |
| Box-grained reranking matching for multi-camera multi-target tracking [13] | 2022 | Cascade-RCNN for detection + DeepSort based framework for SCT + ResNet based deep feature extraction + box-grained cost matrix with spatial-temporal constraints | 0.8486 | HL spatial-temporal constraint |

## 3. Methodology

The general architecture of our proposed Multi-Camera Vehicle Tracking (MCVT) algorithm is depicted in Figure 1 and has five key steps: object detection, feature extraction, SCT, CLM and multi-camera matching. The steps and dataflow are as follows: 1) the object detector obtains the bounding box (BBox) location for each detected object in every frames; 2) the cropped images of these detected objects are then used as inputs for three different ResNets to extract deep appearance features; 3) t using the BBox and Re-ID features, the SCT creates tracklets for each target within a single camera's view; 4) the CLM is trained using the results from the single camera tracking and the ResNets' deep features; 5) the results from single camera tracking and the deep features are used to form a cost-matrix, which is constrained by the spatial-temporal information generated by our CLM, and the matrix is then solved to match track IDs across cameras. More detailed descriptions of these processes are given below.

### 3.1 Object Detection

A reliable vehicle detection process is a prerequisite for vehicle tracking. Many MCMCT problems use YOLO as the object detector [37] due to its accuracy and ability to infer in real-time.. The popularity of YOLO has led to extensive development, the current iteration YOLOv9, considered as the SOTA for its category of detectors. Here, we employ the YOLOv9e model, which is pretrained on the COCO dataset, to specifically detect cars, motorcycles, buses and trucks. To prevent the same target from being recognized multiple times under different categories, we implement non-maximum suppression (NMS) and confidence score filtering across all detected objects. This produces more reliable detection results whilst minimizing redundant detections and mitigating issues caused by occlusions.

### 3.2 Re-ID Feature Extraction

The deep embedding quality makes a significant contribution to the final matching results. Building on [11], we utilize three distinct ResNet models: ResNet50IBN-a, ResNet101-IBN-a, and ResNeXt101-IBN-a. All weights are pre-trained on the CityFlow dataset with a combination of softmax cross-entropy loss function and triplet loss [11]. The cost function $L_{reid}$ is given by

$$L_{reid} = L_{cls} + \alpha L_{trp} \quad (1)$$

where $L_{cls}$ and $L_{trp}$ represent the softmax cross-entropy loss and triplet loss, respectively, with α balancing their weights. Each ResNet model outputs a [2048×1] dimension feature vector and the final feature of each detected car is the average output of the three models.

### 3.3 Single Camera Tracking

In SCT, we follow the track-by-detection approach to associate each detection in the video frame with a synchronized track id. Considering the computational load and accuracy, DeepSort [1] is selected as our main framework. DeepSort integrates a Kalman filter, based on a constant velocity model, to predict locations. This prediction is then merged with deep appearance features to formulate tracklets. One of DeepSort's main advantages is the inclusion of the *Matching Cascade* approach to mitigate the impact of short-term occlusions, which is a critical issue for MCVT problems.

### 3.4 Camera Link Model

The Camera Link Model (CLM) leverages temporal and topological information between different cameras to establish spatial-temporal constraints, thereby enhancing the performance of multi-camera matching. This model includes camera link information and the transition time distributions between adjacent cameras. Our proposed CLM is shown in Figure 2 and provides the exit/entry zone pair of adjacent cameras with a time-transition kernel estimation window via a self-supervised training process. Two cameras are able to form a camera link if they are considered to be adjacent, i.e. there are no other cameras that must be passed when moving between them. Each link will have two zone pairs, one entry/exit pair and one exit/entry, representing the bidirectional flow between two linked cameras.

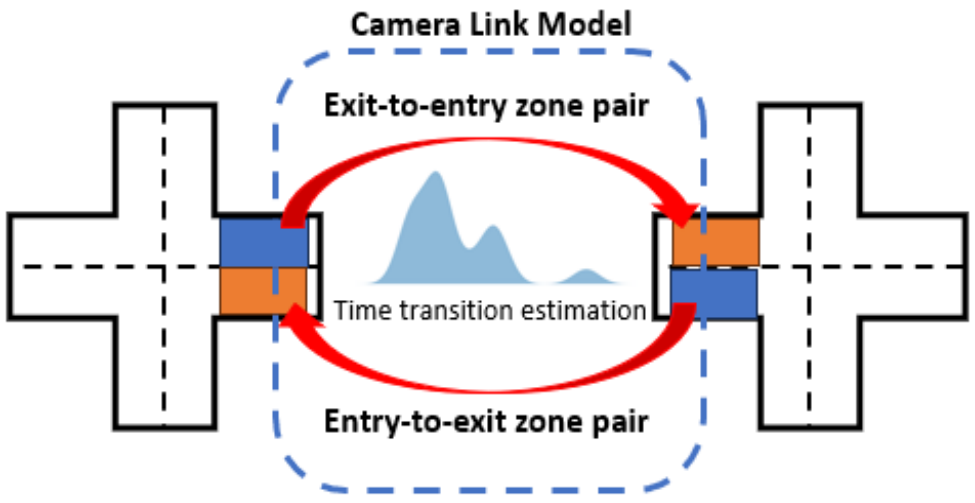

**Figure 2.** CLM illustration. The orange and blue rectangle represent the entry and zone for a certain junction respectively. The time transition estimation is the visualization on kernel density estimated time window.

Based on the camera link, high confidence tracklet matching results, passing through the zone pair, are used to perform the kernel density estimation (KDE), which helps the camera link time transition estimation. The proposed CLM can be divided into three steps: 1) entry/exit zone generation in single camera; 2) entry-exit zone pairing across adjacent camera; 3) camera link transition time estimation.

*3.4.1 Entry/exit zones generation.* Due to road structure and traffic rules, all vehicles must follow one of a set of particular movement patterns. The generation of the entry/exit zones is estimated by utilizing these distinct patterns. We select complete tracklets from the SCT and extract the entry and exit points of each tracklet along with their vector directions to identify consistent movement patterns. In scenes with non-overlapping FOV, where typically only one camera covers an intersection, the challenge of visual perspective arises: vehicles that are closer to the camera appear significantly larger than those further away. This will clutter the entry/exit zones, causing issues for the automatic zone generation. To overcome this, we perform a pre-clustering step based on the direction of travel, before the point distance clustering. The MeanShift algorithm [33] is used for our clustering process. Figure 3 shows an example of the auto-zone generation process and also illustrates the difference between pre-clustering and non pre-clustering.

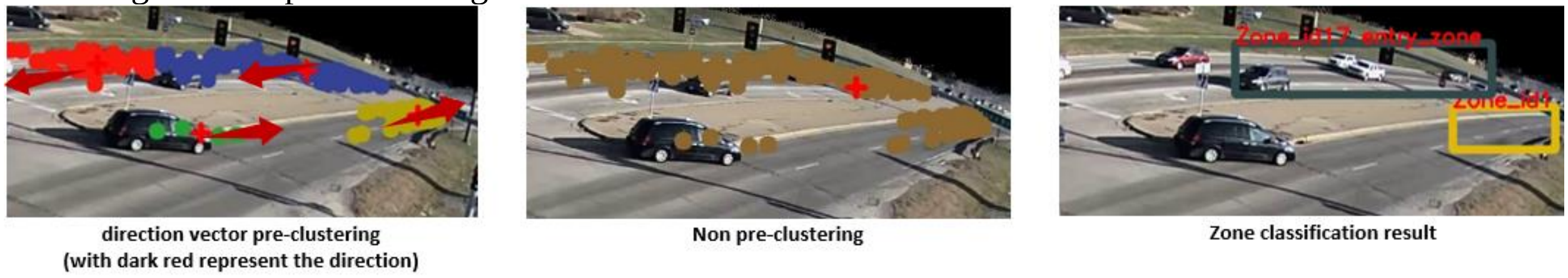

**Figure 3.** Auto zone generation example. The red arrow denotes vehicle moving direction of specific cluster, the red cross marking is the cluster centre, and the rectangle represents the formed zone.

After the clustering process, once the number of points in cluster is higher than $N_{thres}$, the entry and exit densities ($D_e$ and $D_x$) are calculated and used to classify the clusters into different zone classes using:

$$zone\ class = \begin{cases} entry\ zone & if\ D_e > \rho_e, \\ exit\ zone & if\ D_x > \rho_x, \\ undefined\ zone & otherwise. \end{cases} \qquad (2)$$

*where the densities*

$$D_e = \frac{N_{e,k}}{N_{e,k}+N_{x,k}}, D_x = \frac{N_{x,k}}{N_{e,k}+N_{x,k}} \qquad (2)$$

in which $N_{e,k}$ and $N_{x,k}$ are the number of entry points and exit points, respectively, in cluster $k$. If the entry/exit density is higher than a threshold, $\rho$,, a rectangle window that encompasses all points inside this cluster is defined as the corresponding entry/exit zone.

*3.4.2 Entry-exit zone pairing across adjacent camera.* Following automatic zone generation, camera links are established by defining entry-exit zone pairs across adjacent cameras. Each entry-exit zone pair represents the path a vehicle takes when it exits from one camera's zone and enters another camera's zone, without deviating onto side roads. This ensures that the trajectory of the vehicle is continuously tracked across multiple camera views. To find the entry-exit zone pair between adjacent cameras, a set of high confidence tracklets are defined as those exceeding pre-defined thresholds for distance and duration. These tracklets are then compiled into a potential zone matching database. A pre-tracklet matching for each potential zone pair is conducted using the same approach as in multi-camera matching, but without incorporating spatial-temporal constraints. This cost-matrix formation and solving procedure will be explained in detail in the later section.

To assess the confidence of each zone pair, we introduce a zone pair confidence score calculated from the preliminary tracklet matching results. This zone pair confidence score consists of three parts 1) the average cosine distance between tracklet pairs, 2) the number of matched high confidence tracklets, and 3) the time variance of these matched tracklet pairs. This zone pair confidence score function is formed as:

$$Score = -\alpha Distance_{cosine} + \beta N_{pair} - \gamma Var_{time} \tag{4}$$

where, $\alpha$, $\beta$, and $\gamma$ are coefficients that weigh the importance of each component, $Distance_{cosine}$ is the average cosine distance for paired tracklets, $N_{pair}$ is the number of tracklet pairs and $Var_{time}$ represents the variance in transition time for each pre-matched tracklet pair. Considering our test dataset road topology, each camera connects with one adjacent camera, the calculation process follows the sequence [c041,c042], [c042,c043] … [c045,c046] and the zone pair corresponding to the highest confidence scores are selected to be the entry/exit zone pair. This method ensures that the most reliable pathways between cameras are identified based on quantifiable metrics, facilitating accurate vehicle tracking across camera networks.

*3.4.3 Camera link transition time estimation.* To estimate the transition time of each camera link, the high confidence tracklet pre-matching results for paired zones are used. The transition time for paired tracklets is calculated by:

$$T_{tran} = T_{exit} - T_{entry} \tag{5}$$

We then filter out any negative transition times ($T_{tran} < 0$), as these are not feasible. The remaining $T_{tran}$ is used by a Gaussian kernel density estimation (KDE) to find the time transition estimation between two cameras. Comparing with the set hard-removal time-window between linked cameras approach, the KDE approach aligns the transition possibility for each transition time, contributing to a more accurate method for identifying the potential pair across different cameras. This KDE result provides hard temporal removal constraints (if the possibility is too low) while also being able to contribute to the cost matrix matching process.

### 3.5 Multi-Camera Matching

The temporal and spatial information provided by our CLM provides the inputs to the matching process. The multi-camera matching mainly relies on solving a cost matrix of tracklets across different cameras. Our cost matrix consists of: 1) deep appearance distance; 2) transition time estimation; 3) spatial-temporal information mask. The first step is to calculate the distance between each tracklet's appearance vector. This appearance vectors derived from the object Re-ID feature that is processed by the temporal attention mechanism [34] to generate a 2048x1 dimension feature. The deep appearance distance of tracklets $T_i$ and $T_j$ can be computed using cosine similarity distance:

$$Dis_{cos}(T_i, T_j) = 1 - \frac{F(T_j) \times F(T_i)}{|F(T_j)| \times |F(T_i)|} \tag{6}$$

where $F(T_j)$ is the feature of tracklet $T_j$ and $F(T_i)$ is the feature of tracklet $T_i$. In addition to the cosine distance, the transition time estimation is taken into account. Combing those two parameters together, the cost distance is formed:

$$Cost(T_i, T_j) = \delta Dis_{cos}(T_i, T_j) - \epsilon KDE(T_{tran}(T_i, T_j)) \tag{7}$$

where, $\delta$ and $\epsilon$ are coefficient weights for the appearance and transition time factors in the cost function, $KDE$ represents the kernel density score for the transition time. A higher KDE score indicates a more likely transition between the cameras. From all those above we can get the cost matrix C for $m$ tracklets as:

$$C = \begin{bmatrix} Cost(T_1, T_1) & \dots & Cost(T_1, T_m) \\ \vdots & \ddots & \vdots \\ Cost(T_m, T_1) & \dots & Cost(T_m, T_m) \end{bmatrix} \tag{8}$$

After forming the initial cost matrix, a spatial-temporal mask inspired by [11] is used to filter unlikely matching pairs that reduce the searching space. For tracklet $T_i$ and $T_j$, we decide whether they conflict with each other using the rule base defined in Table 2.

**Table 2.** Spatial-temporal confliction rule.

| tracklet in query camera | tracklet in gallery camera | Time | Conflict |
|---|---|---|---|
| paired entry zone | paired exit zone | $KDE(T_{tran}) < \rho$ | TRUE |
| paired entry zone | paired entry zone | Any $T_{tran}$ | TRUE |
| paired exit zone | paired entry zone | $KDE(T_{tran}) < \rho$ | TRUE |
| paired exit zone | paired exit zone | Any $T_{tran}$ | TRUE |
| Unpaired entry and exit zone | Any tracklets | Any $T_{tran}$ | TRUE |
| Any tracklets | Unpaired entry and exit zone | Any $T_{tran}$ | TRUE |

Table 2 lists all situations that two tracklets conflict with each other. If a tracklet entry or exit in the entry/exit zone it will be assigned into the corresponding entry/exit zone dataset, e.g. one tracklet entry the camera c041 scene in zone 1 and exit in zone 7, it will be assigned into the zone 1 dataset and zone 7 dataset. For the adjacent camera pair, there are two entry/exit zone pair which are defined as the paired entry/exit zone for building the confliction rules in Table 2. Referring to this spatial-temporal confliction rule, it will form a mask matrix:

$$M = \begin{bmatrix} Mask(T_1, T_1) & \dots & Mask(T_1, T_m) \\ \vdots & \ddots & \vdots \\ Mask(T_m, T_1) & \dots & Mask(T_m, T_m) \end{bmatrix} \tag{9}$$

where

$$Mask(T_i, T_j) = \begin{cases} True & if\ Conflict\ = True \\ False & elseif \end{cases} \tag{10}$$

Finally, we combine the cost matrix with our spatial-temporal mask:

$$\hat{C} = C \odot M \tag{11}$$

where $\odot$ represents the elemental-wise product of the corresponding elements of the matrix. The constrained cost matrix given by (9)is used for subsequent tracklets clustering. After filtering through the spatial-temporal mask, the searching space is significantly narrowed. In this case, we greedily select the smallest pair-wise distance to match the tracked vehicles. For each ordered transition, we further remove the pairs which conflict with previously matched pairs and pairs that have distance higher than threshold $SychThresh$. We repeat the process until there is no valid transition pair or the minimum distance is larger than a threshold.

## 4. Experiments and Results

### 4.1 Dataset

CityFlow [25] is the most representative and the largest MCVT dataset that captured in the actual scene of the city. The dataset includes at least 3.25 hours of traffic video at 960p resolution from 40 cameras across 10 intersections in a medium-sized city, covering a total length of approximately 2.5 kilometers. It features a variety of road camera setting topologies, including non-overlapping, mixed, and overlapping FOV. The test set consists of 6 cameras with non-overlapping FOV, which we use to validate our proposed algorithm.

### 4.2 Evaluation Metric

For the MCMOT problem, many evaluation metrics are used, including but not limited to: MOTA, IDF1 and HOTA . In our task, we chose the IDF1 score to be the metric to evaluate the performance of our proposed algorithm. IDF1 calculates the ratio of the number of correctly identified detections to the ground truth and the average number of calculated detections.

$$IDF1 = \frac{2IDTP}{2IDTP + IDFP + IDFN} \tag{12}$$

where IDTP, IDFP, and IDFN represent the counts of true positive, false positive, and false negative identifications, respectively.

### 4.3 Implementation Details

Our experiments are implemented on our own PC installed Ubuntu 20.04.4 LTS OS with 13th Gen Intel(R) i7-13700KF CPU and NVDIA GeForce RTX 4070 Ti GPU. For object detection, the YOLOv9e model pre-trained on coco dataset is applied. The detection confidence score threshold is set to be 0.2, with NMS-IOU set 0.45. In the feature extractor training process, we respectively use ResNet50-IBN-a, ResNet101-IBN-a and ResNeXt101-IBN-a as the backbone for training and those trained networks are used to extract $2048 \times 1$ dimension deep feature. In SCT, the DeepSort-based framework was employed, with a minimum confidence score of 0.2 and a minimum IOU of 0.5 between predicted and current BBoxes. For the camera link mode, the cluster threshold $N_{thres} = 5$, the threshold for entry/exit zone definition $\rho_e$ *and* $\rho_x$ is 0.8. The zone pair confidence score weights $\alpha, \beta, \gamma$ are set to be 0.7, 0.3 and 0.1, respectively. For multi-camera matching, $\delta$ and $\epsilon$ are set to be 1, -0.5 and the KDE filter $\rho$ is set to be 0.001. All hyperparameters above are experimentally found to give the best IDF1 performance. In the end, total 170 tracks across multiple cameras in our test set are detected, and the resulting IDF1 score of 0.6107 surpassed all other camera link-based approaches.

### 4.4 Ablation Study

In this section, an ablation study is conducted on the CityFlow V2 test dataset to determine the individual contributions of different modules within our proposed framework. This study helps to isolate the effects of each component on the overall performance, as detailed in Table 3.

**Table 3.** MCVT ablation study. Various algorithms are replaced / added on a baseline method (YOLOv8 + average mean feature extraction + DeepSort + Greedy Matching) to evaluate the performance improvement on those modules.

| Method | IDF1 | IDP | IDR | Precision | Recall |
|---|---|---|---|---|---|
| Baseline | 0.4125 | 0.4807 | 0.3612 | 0.5189 | 0.39 |
| + Parameter Fine Tuning | 0.4462 | 0.4791 | 0.4175 | 0.5197 | 0.4528 |
| + Yolov9e | 0.4524 | 0.4988 | 0.4139 | 0.5396 | 0.4477 |
| + Cosine Similarity | 0.4686 | 0.5063 | 0.4362 | 0.5432 | 0.4679 |
| + Temporal Attention | 0.485 | 0.5273 | 0.449 | 0.5556 | 0.4731 |
| + Spatial Constraint (CLM) | 0.5733 | 0.6248 | 0.5296 | 0.6512 | 0.552 |
| + Temporal Constraint (CLM) | 0.6107 | 0.6471 | 0.5782 | 0.6636 | 0.593 |

In this ablation study, a baseline method is conducted as the YOLOv8x + 3 ResNets combined feature embedding + DeepSort SCT + Greedy Matching without any constraints setting. The study evaluates the impact of tuning various modules on the system's performance. However, the primary highlight is the substantial impact of the proposed CLM, which provides both spatial and temporal constraints, on enhancing the tracking performance by 0.1257 IDF1 score.

### 4.5 Result Analysis

Based on the ablation study, we prove the improvement provided by our improved CLM. However, comparing the performance improvement with human-annotation approach [11], our CLM still does not leverage all the potentials for spatial-temporal constraints. A primary challenge identified is the imperfect zoning process, which can be illustrated by the zone generation example in Figure 4.

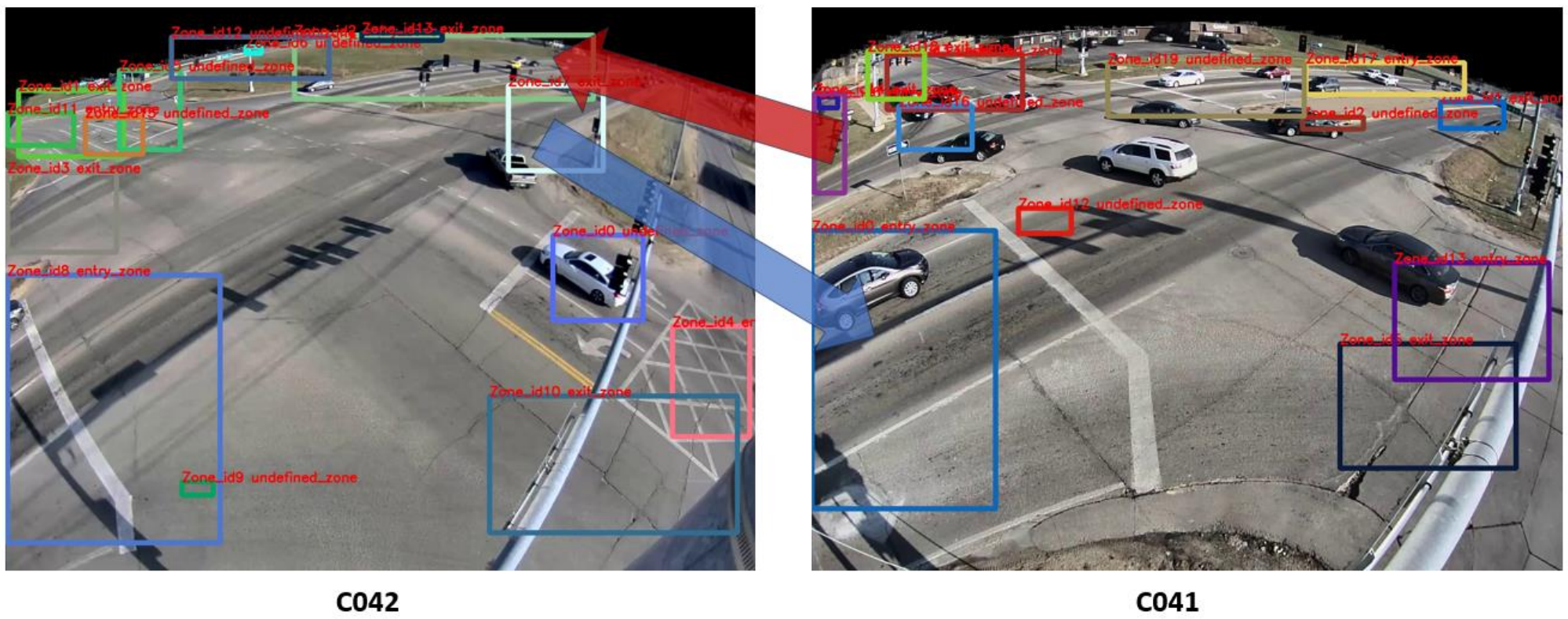

**Figure 4.** Cross camera zone generation and pairing example

In the example shown in Figure 4, the clustered zone is formed in certain areas but does not completely align with the actual exit/entry points based on the real topology of the road. while also leaving several areas as undefined zones, even though tracking is performed in those entry/exit zones. Despite the zoning issues, cross camera zone-pair achieves 100% accuracy in the test dataset which contains 6 cameras. In addition, the time transition kernel density estimation results in a 0.01 increase in IDF1 score on the test set compared with the human-annotated hard time window constraints. Overall, the improved CLM provides valuable improvement on the overall tracking performance based on IDF1 score evaluation metric. However, the limitation of the dataset size and some potential flaws on our algorithm prevent us from leveraging the entire potential of the spatial-temporal constraints compared with human-labelled constraints. A bigger dataset (longer video footage for each camera) might lead to better training for our CLM, which could potentially reveal further benefits of our approach.

Apart from the CLM, the rest of the system follows the basic framework of current SOTA [12] which holds the second highest IDF1 score. The difference is our approach removes the feature dropout filter, multi-target multi-level association and the multi-tracker combined SCT framework, which brings too much computational load that prevent any potential future real-world real-time deployment application. The removed sections cause our multi camera matching process to suffer from the detrimental impact of false positives and fragmented tracks which leaves the gap for our approach compared with the SOTA.

## 5. Conclusion

In this paper, we present a MCVT framework that incorporates improved CLM. Building on mature methodologies for object detection, feature extraction and SCT, this method introduces novel advancements in spatial-temporal constraints and multi-camera matching modules. The proposed CLM is capable of generating spatial-temporal constraints autonomously, eliminating the need for human intervention. Our ablation study validates the effectiveness of these spatial-temporal constraints within the multi-camera matching process. The innovative approach achieves the 0.6107 IDF1 score, surpassing all other camera link model-based methods and indicating the potential of our proposed framework to provide reliable spatial-temporal constraints. However, our approach still exhibits a performance gap with human annotated SOTA due to imperfect zoning and intentional removal of some computationally intensive modules. This removal, while limiting the IDF1 performance, leaves potential for future real-world online multi camera tracking deployment on our industrial partner's real-world online SAE vision tracking pipeline [38]. This vision tracking pipeline provides a framework for future MTMCT tracking tasks not just limited on vehicles but also on people.